\documentclass[journal]{IEEEtran}
\ifCLASSINFOpdf
\else
\fi
\usepackage{url}

\usepackage{graphicx}
\usepackage{amsmath}
\usepackage{amsfonts}
\usepackage{amsthm}
\usepackage{tabularx}

\usepackage{tabularx}

\begin{document}
%
\title{Structured Memory based Deep Model to Detect as well as Characterize Novel Inputs}
%
%
%

\author{Pratik~Prabhanjan~Brahma, Qiuyuan Huang,
        Dapeng~Wu,~\IEEEmembership{Fellow,~IEEE}
\thanks{P.P. Brahma is now with the Volkswagen Group of America Electronics Research Lab, Belmont, CA 94403. Q. Huang is now with the Microsoft Research, Redmond WA. However, this work was done while P.P. Brahma and Q. Huang were graduate students at University of Florida, Gainesville, FL 32611 USA.

D. Wu (correspondence author) is with the Department
of Electrical and Computer Engineering, University of Florida, Gainesville, FL 32611, USA. Corresponding Email: dpwu@ufl.edu}
}

\maketitle

\begin{abstract}
While deep learning has pushed the boundaries in various
machine learning tasks, the current models are still far away from
replicating many functions that a normal human brain can do.
Explicit memorization based deep architecture have been recently proposed
with the objective to understand and predict better. In this work, we
design a system that involves a primary learner and an adjacent representational memory bank which is organized using a comparative learner. This spatially forked deep architecture with a
structured memory can simultaneously predict and reason about
the nature of an input, which may even belong to a category never seen in the
training data, by relating it with the memorized past
representations at the higher layers. Characterizing images of
unseen object classes in both synthetic and real world datasets is used as an example to showcase the
operational success of the proposed framework.
\end{abstract}


%
\IEEEpeerreviewmaketitle

\section{Introduction}
%
%
%
%
\IEEEPARstart{A}{rtificial} Intelligence (AI) and more
specifically deep learning research have been moving in the
direction of emulating the human brain both functionally as well
as structurally.
A human brain is capable of learning, understanding,
discriminating, imagining, inferring, memorizing, retrieving, inventing,
providing feedback and deciding how to act in a given situation.
Although the various components of the human brain perform
differently, the fundamental building units are based on neurons
and synapses. As has been common in deep learning architecture
\cite{DLbook}, neurons are also connected in hierarchical layers
in order to capture various levels of semantic and abstract
information about a given input. It is encouraging to see that the
modern neural network models can provide visually interpretable
insights \cite{zeiler2014visualizing} to support such hypotheses
regarding the human brain structure. The traditional connectionist
approach tries to learn a computational mapping from a vectorial
input to another vectorial output using interconnections of
several units. On the other hand, symbolic logic or reasoning
\cite{connection} involves rule based expert systems, for example
in structured grammar or language models. An AI system should ideally be able
to incorporate both just like the human brain. Here, we present a
framework for deep learning which can not only learn to map input
to output but also learn comparative relationships between input
classes with the help of an explicit memory. Such a model can not
only recognize new samples of the past types but also identify and
characterize novel observations. This would enable an artificial
model to act and think like a
human. This work aims at building a single architecture that is able to act on instances of known types while also be able to think and relatively place novel types using a structure memory framework.   

The concept of associated memory has been the crux of neural
networks where the values of the connection weights store information regarding the training data. Long Short-Term Memory (LSTM)
\cite{hochreiter1997long} try to capture information over time while learning the mapping between input and output. However recent
research, like memory networks
\cite{weston2014memory} and neural Turing machines \cite{graves2014neural}, have been in the direction of explicitly
dedicating resources for storing and retrieving information
regarding the input in a concise yet effective way. These works are motivated towards the use of readable and writeable external memory. This is
analogous to the structure of the brain where neocortex plays the
role of the intelligent learner \cite{hawkins2007intelligence} and analyzer of sensory inputs
whereas thalamus and hippocampus share the responsibilities of
storing short term and long term memories \cite{winocur1985hippocampus}. However, some memories
may be event or pattern based while some may also be relational or
comparative. Some of the examples can be the relationship between
two individuals or color intensity of one flower as compared to
another. In order to learn and memorize such things, one needs to
form relationships with respect to past memories. It is also important to note that
humans tend not to memorize everything that they observe nor do they perform an exhaustive search on all past memories to retrieve or recollect something. The search and understand mechanism is often instantaneous and is probably due to the format in which memories are stored.  Data structures in computer science have many forms like graphs, lists or trees depending on particular use cases. This motivates us towards coming up with
a structured memory framework. It is also
argued here that this can help learn the true concept behind a
given task and can thus lead to intelligent analysis on even
unseen observations. The next section discusses how our work is inspired yet different from
some of the prior approaches for solving zero shot learning and learning relative attributes. After that, the model design is described along with experiments on synthetic and real world object recognition and characterization tasks.


\section{Related Work}

There has been a long history of research in the field of novelty detection and learning. The closest area of research to our work is Zero shot learning \cite{palatucci2009zero} or zero data learning  where the objective is to find
a generalized model that can infer input belonging to those categories for which no training
data are available. Usually, the only extra information available is some linguistic meaning or short description of the categories which can be used to produce embedding vectors.  Most of the works in this field
have been towards domain adaptation and transfer learning
\cite{DLbook}, i.e.,  a model which has been learned in one
setting or distribution is used to test on another unknown distribution or to perform
a new task. For example, \cite{zdl_aaai} tests the ability of a
character image classifier to discriminate between two
character classes which do not exist in the training set. So, even if the
test domain and its distribution are different, the task is
related or similar for which a generalized model is expected to
serve the purpose.

Many recent research works in this area, for example as in \cite{zero_crossmodal}, use
the concept of category word embedding to detect unseen classes in the low-dimensional semantic word vector space. In the case of a novel image, it classifies the representation of the image with the help of
unsupervised semantic word vectors. However, it requires multi-modal data to map
correspondences between image feature space and word space
vectors and we may not have the luxury of such information is all learning tasks. Also, embedding vectors not resilient to data shift or domain shift. On the other hand,\cite{zdl_aaai} shows how to cluster together inputs from unseen classes.  But such
an architecture is not able to semantically explain a whole new example
 and its unknown class. In \cite{lampert}, the authors introduce an
attribute based classification where object detection is performed
based on human specified high level description of the target
objects. The description can comprise of shape, color or some
other information and new classes can be detected based on their
attribute representation. However, in real life data, we do not
have such rich and fine-grained information for all training images. Also, humans
are still able to judge a novel input without explicitly computing specific handcrafted features or statistics. The theory behind graph based semi-supervised
learning is used in \cite{transductive} to do knowledge transfer 
 for figuring out zero shot classes in datasets like Animals with
Attributes and ImageNET \cite{ILSVRC15}. They also use external linguistic
knowledge and similarity measures to do such learning. On
the other hand, a semantic output code
based classifier is proposed in \cite{Hinton_zero}  and its application is shown on predicting words
that people are thinking about by using corresponding fMRI images without training
examples for those words. This is yet another example that involves using
multi-modal data. On the similar path, one shot
learning \cite{feifei06} is when only one (or very few) of the
instances of that category is shown to the model during the
training phase. The motivation behind such work is the ability of humans to understand and recognize objects which they might have seen only once. However, such works have not leveraged the mutual
relationship between classes which may come in need during certain
problems. They have also not shown how the memory of a novel class
can help reshape the knowledge of the past comprising of inputs
belonging to known categories.

In the absence of embedding meta information regarding each class, relative attributes  between them can also help in zero shot recognition. For example, \cite{parikh2011relative} trained a binary support vector classifier to learn the relative attribute given a pair of images. Deep learning has also helped in raising the bar, as shown in \cite{souri2016deep} where a Siamese style network \cite{chopra2005learning} with shared weights followed by a ranking layer was used to predict the relationship between a pair of given images coming from datasets like the UT-Zap50K \cite{finegrained}. However, these kind of approaches do not provide any single end-to-end model for recognition, detection and characterization of instances from both known and unknown categories. Also, training the whole network with all pairs of images from a dataset demands a lot of time and storage requirement. Although our proposed architecture is inspired from the work done in the field of learning relative attributes, we compare instances at the feature level rather than at the pixel level. Thus, our model has fewer trainable parameters and can also be placed along with a stand alone deep classifier. The detailed architecture is discussed in the next section. Moreover, the relationship between categories, or training examples, can take any real values and not just binary. Given the speed at which humans recall and realize, it should be possible to have an end-to-end system that can learn to recognize while also being able to understand newer examples based on past memory. We especially take inspiration from nature where it is the combined effort of neocortex's intelligence and hippocampus's dedicated memory that helps in learning things faster and retain the knowledge. The memory is formed using representations and it is organized as a weighted graph, or correspondingly an adjacency matrix, where the edge weights correspond to the relative attribute information which can take any real values. In this paper, we motivate and
experiment on a problem which requires such relational
intelligence and propose a structured memory learning based deep
architecture to deal with it.


\section{System Design}

\begin{figure}[htbp] 
\includegraphics[width=3.5in]{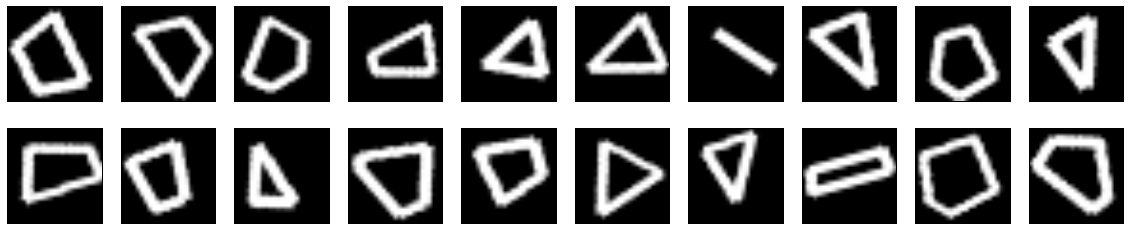}
\caption{Images of four known geometrical shapes present in the training dataset}
\label{poly}
\end{figure}

\begin{figure}[htbp] 
\includegraphics[width=3.5in]{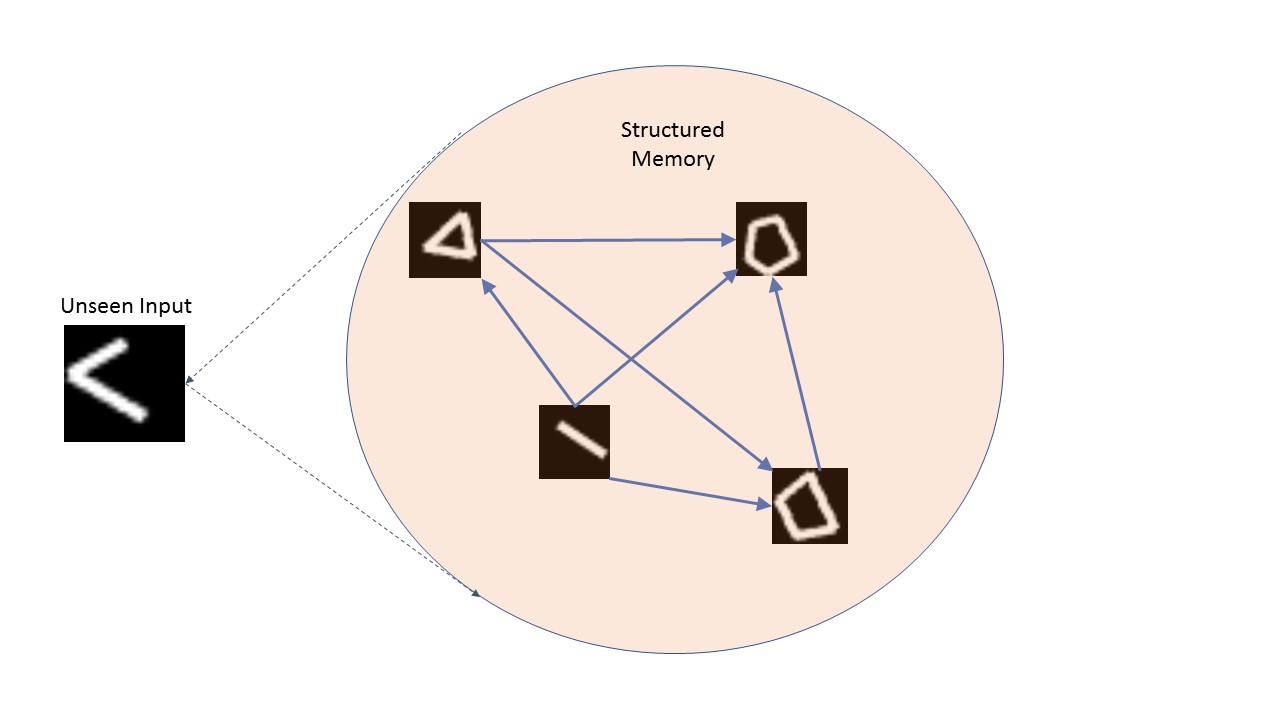}
\caption{Input from an unseen class (angles) is tested against members of a structured memory representation bank. Only positive edges are shown. For each edge, there exists a negative edge in the reverse direction. Self connections and no connections have zero weight.}
\label{struct}
\end{figure}

In the case of object recognition, most neural networks project the input data onto some representation space where a simple classifier can do accurate recognition. However, a representative feature space can also build the base for creative understanding of the data rather than simply recognizing it. For example, the task in Fig. \ref{poly} is to classify gray scale images of geometrical
shapes.  The classes are defined by the number of edges in it. That is exactly the
concept that needs to be understood by a student to be able to compare a completely
new geometrical shape with respect to the ones seen before.
However, a conventional discriminative network will either
wrongly classify it as one of the known shapes or predict it as an
outlying polygon. Zero shot models like \cite{zdl_aaai} may transfer a trained network to also do a
fine job in distinguishing an unseen hexagon from an unseen
heptagon  but  won't be able to reason why or even characterize a
single hexagon image. Our proposed model explicitly memorizes instances from the
past along with learning the true concept in order to characterize novelty while also recognizing old ones.
This will help a model emulate both acting and thinking like humans. In addition to primarily approximating the mapping from the images
to its output class using a neural network (which we call as the primary learner), the architecture also has a comparator module
that can figure out the relationship between representations of each
class. The comparator can have various structures depending on the concerned task. But most importantly, the artificial system has to learn the true concept
(not explicitly mentioned during training) that can not only
discriminate between the known classes but also be able to characterize
novel ones.

The input data matrix
$\mathbf{X}$ consists of $N$ exemplars or images. The target $\mathbf{Y}$ is a $N \times 1$ vector that
denotes the corresponding labels. Along
with $Y_i$ for every $X_i$, we also have the comparator target
matrix $\mathbf{Z}$ whose $(i,j)^{th}$ entry is the answer to a
conceptual question given a pair of inputs.  Now, when a new input is fed into the network, it is redundant to
compare it with each and every past instances seen in the training
data. That is where the idea of having sampled representatives
comes into picture. We rather treat the representation space at
higher layers \cite{brahma2015deep} as a disentangled and
invariant feature space and memorize samples
 which are needed to train the comparative learner whose target values are given by $\mathbb{Z}$.
This way, the number of instance comparisons for training the
comparator would be far less, of $\mathbf{O}(C^2)$ instead of
being $\mathbf{O}(N^2)$ where $C$ and $N$ are the number of categories and training examples respectively.  The primary learner can be either a
discriminative or a generative neural network with bottleneck layers as we are mostly
interested in compressed yet informative representation space for
the input data. For our current experiments, we show that discriminative neural architectures commonly used  for object recognition can fit well as a primary learner. The primary learner is trained using the loss function $l_P(\mathbf{\Theta})$,
 where $\mathbf{\Theta}=[\Theta_1,\Theta_2...\Theta_d]$ represents the weights and biases of the network with depth $d$.
 At the $i^{th}$ layer, representations can be sampled to form the representation bank $\mathbf{R}_i \in \mathbb{R}^{M \times l_i}$.
 It is important to note that the number of sampled memories, $M$, is a constant number and much less than the total number of training observations $N$. The combined result of all the transfer functions $f_i(.)$, which refers to the activation function of the $i^{th}$ layer, and the sampling function $g(.)$ at the higher layer gives rise to the entities of the representation memory bank

\begin{equation}
\mathbf{R}_k^{i}=g(f_i(...f_2(f_1(X_k))))
\end{equation}

\begin{figure}[htbp] 
\includegraphics[width=4.5in]{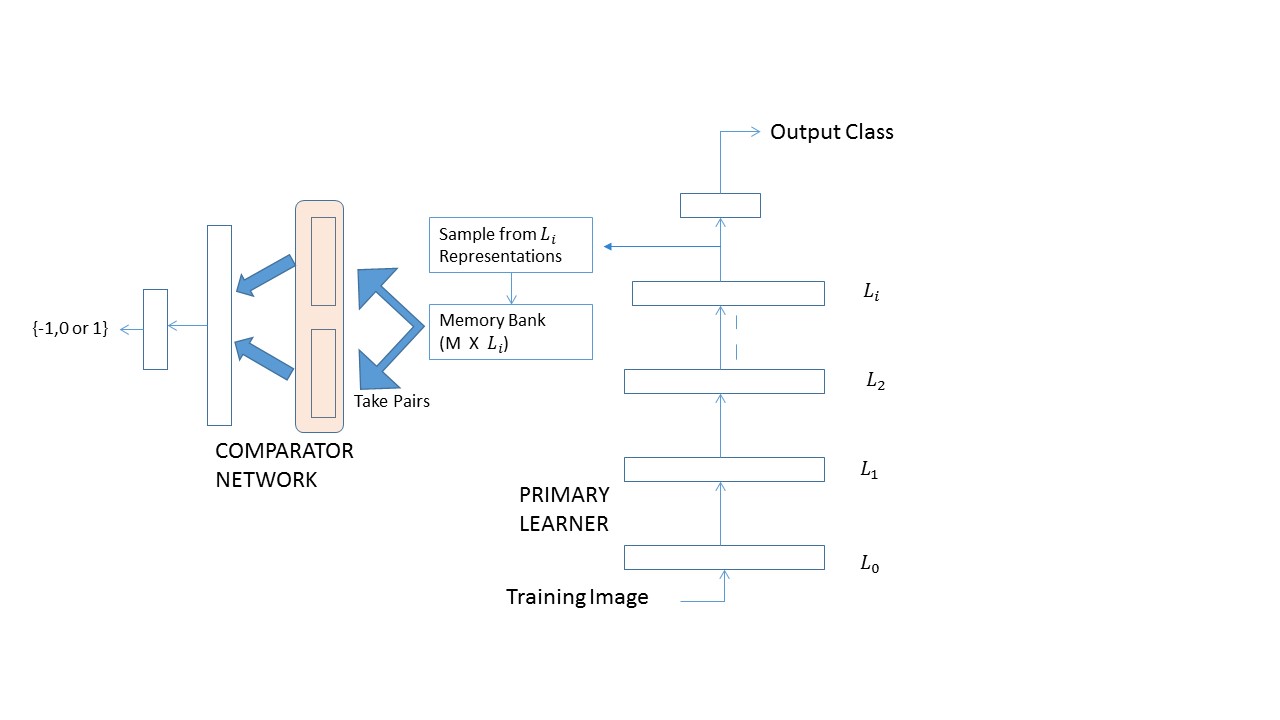}
\caption{A primary learner and a comparator network working on sampled representations at higher layers during training phase. Primary Learner performs the classification whereas comparator takes pair of representations and outputs a relative attribute. }
\label{train}
\end{figure}

\begin{figure}[htbp] 
\includegraphics[width=4.5in]{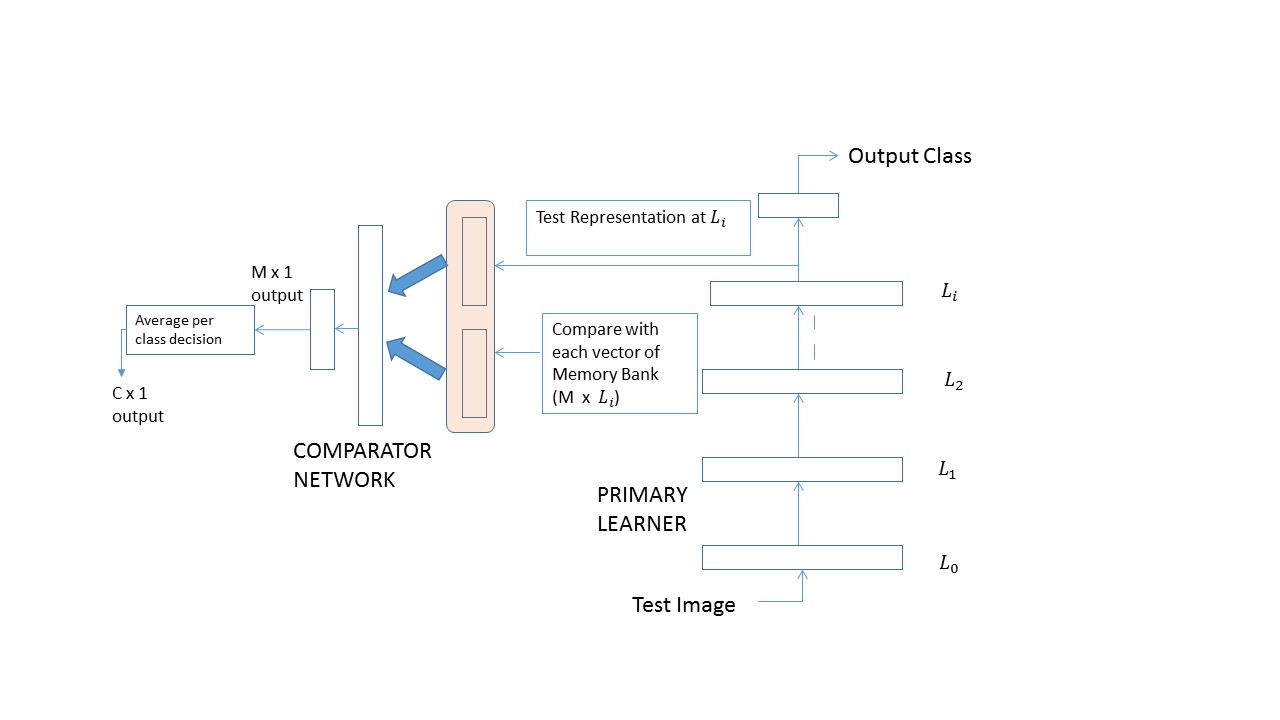}
\caption{The inference phase involves the primary learner performing classification and detecting instances where it is not confident. In the latter case, the input's corresponding representation is paired with past feature memories from representation bank and fed to the comparator network to make a category wise decision.}
\label{decide}
\end{figure}

Choosing the right sampling function, $g(.)$, can be framed as a subset selection problem and hence submodularity \cite{submodular1} can be used to select the samples that best represents the dataset. One way to solve it is using the k-medoids approach where a subset $S$ of $k$ centrally located exemplars are chosen that minimize the sum of dissimilarities $d(.,.)$ between every data point and its cluster center

\begin{equation}
S^{*}=\text{argmin}_{|S|\leq k} L(S)=\text{argmin}_{|S|\leq k} \frac{1}{|V|}\sum_{v \in V}{\text{min}_{e \in S} d(e,v)}
\end{equation}

Submodular set function has the property that the difference in its incremental value that a single element makes when added to an input set decreases as the size of the input set increases. Mathematically, a set function $f:2^V-\mathbb{R}$ over the whole dataset of $V$ is considered to monotonically submodular if for every $A \subseteq B \subseteq V$ and $e \in V - B$

\begin{equation}
f(A \cup {e}) - f(A) \geq f(B \cup {e}) - f(B)
\end{equation}
The objective function of k-medoids can be framed as a  monotone submodular function \cite{submodular2} by introducing an auxiliary element $e_0$ as
\begin{equation}
f(S)=L({e_0})- L(S\cup {e_0})
\end{equation}
where $f$ measures the decrease in the loss associated with
the set $S$ versus the loss associated with just the auxiliary
element $e_0$. Maximizing the submodular function is equivalent to minimize $L(S)$ and hence can be achieved greedily. Facility location and other kinds of submodular functions can also be applied to come up with a representative subset. A faster way is to just do linearly complex Lloyd's algorithm based k-means clustering on the deep feature outputs of each known categories separately and appoint the nearest data point to cluster center as corresponding class representative into a representation bank $\mathbf{R}$. The next step is to give structure to this representative memories using the comparative learner.

We employ Siamese networks \cite{chopra2005learning} with two input chains and one output for the comparator network. This is used as the adjacent structured memory based architecture to learn the relationships between the memories sampled in the training phase.This can be a fairly shallow network with fewer layers since we already have attained information from the input which are invariant to nominal changes. During the training phase, the input to the comparator model is pairs of tensors from the representation bank $(\mathbf{R}_i^{c_a},\mathbf{R}_i^{c_b})$ from classes $c_a$ and $c_b$. If the training of the comparator network is performed independent of the primary learner by using offline stored representation bank, then the loss function here is $l_C(\Gamma_1, \Gamma_2,...\Gamma_q)$. But we can also simultaneously optimize the loss functions of both the primary and the comparative learner as 

\begin{equation}
l_P(\Theta_1,\Theta_2...\Theta_d)+\lambda l_C(\Theta_1,\Theta_2...\Theta_i,\Gamma_1, \Gamma_2,...\Gamma_q)
\end{equation}
 Fig.~\ref{train} details the entire training process of our system.

 During the testing phase, each new image shall output a single class identification as per the primary learner. In addition to that, its corresponding representation will be compared with the $M$ sampled memories in the representation bank to output equal number of predictions to indicate what those memories, as per each previously known category, think about this new input. As can be seen from Fig. \ref{struct}, the relationships in the case of understanding images of geometric shapes can take values from a ternary set ${-1,0,1}$ . Thus, it is asked to learn a skew symmetric non-linear mapping function. 

\begin{equation}
t(\mathbf{R}_i^{p},\mathbf{R}_i^{q})=-t(\mathbf{R}_i^{q},\mathbf{R}_i^{p})
\end{equation}
  However, relative attributes need not be just binary or ternary. They may well be real valued scalars or even vectors with multiple values for which the comparative learner is trained as a regression task. In that case, the decision process can be made dependent on a threshold value $\gamma$ which is basically the extent of influence for each of the possible outcomes. Assuming we have $\eta_k$ number of representatives in the representation bank for category $k$ (with $E[\eta_k]=M/C$ assuming examples are uniformly distributed among all categories), the decision of a new test input $X_{test}$ as perceived by the representatives of each category $k$ can be given as

\begin{equation}
\mathbf{sgn}(\frac{1}{\eta_k}\sum_{Y_j=k}{t(\mathbf{R}_j,\mathbf{R}_{X_{test}}})
- \gamma)
\end{equation}
  Here, the threshold $\gamma$ is a hyperparameter. Using a combination of the $C \times 1$ output vector, we can make a judgment of whether this is a new instance of a previously known category or a whole new unseen object class altogether. The comparative learner can also be trained as a classifier and we did experiments with both approaches in the next section. Fig.~\ref{decide} explains how the model functions during this phase.


\section{Experimental Results}

\subsection{Synthetic Geometrical Shapes}
Here, we first train the model with the categories of line segments,
triangles, quadrilaterals and pentagon as shown in Fig. \ref{poly}
so that after training, an image containing an angle would be a
zero shot class. Thus, the number of seen classes $C=4$. The relative concept that the structured memory needs to learn is -\emph{Given a pair of inputs, does } $X_i$ \emph{have greater (output
+1), less (output -1) or equal number of edges (output 0) as that
of $X_j$ ?}   The geometric shape characterization problem, just like any other similar problem, can be broken down into the following:
\begin{enumerate}
\item
$\mathsf{Concept}$- Number of straight edges it contains. \
\item
$\mathsf{Learning Problem}$- Identifying the type given an image
\item
$\mathsf{Reasoning}$- How complex is one compared to the other?
\end{enumerate}

The experiments were conducted using a Convolutional Neural
Network (ConvNet) and multi-layered Deep Neural Network (DNN) as the
primary learner. The learning modules were written  using GPU
compatible python based deep learning package Keras \cite{Keras}. The ConvNet had
$\mathtt{Conv0+Conv1+MxPool1+Conv2}$ $\mathtt{+Conv3+MaxPool2+Flatten0+FC1+FC2+SoftMax}$ layers and the DNN has multiple hidden layers with effectively the same number of trainable parameters. Dropout
layers and $l_2$ weight penalty on weights were applied to both to
achieve suitable regularization. $N=30000$ number of $25 \times
25$ pixels images were artificially generated in a balanced manner
to $C=4$ number of classes (line segments, triangles,
quadrilaterals and pentagons). Images of angles were provided to the network as novel unseen class patterns during the testing phase. Various factors of variation
including translation, rotation, intensity, scaling and affine
modification were considered while preparing the dataset. As
expected, ConvNet performed better in the classification task  with an
accuracy of 98.25$\%$ whereas the DNN could only go up to
96.63$\%$ on a test set of $5000$ images belonging to the known
$4$ classes. The comparator network was just a shallow two hidden
layer neural network on top of the sampled representations from
the penultimate fully connected layer in both the ConvNet and DNN.
With $M=100$, we had 10000 pairs to train the comparator. However, it's interesting to note that 
similar accuracy could also be achieved by randomly ignoring about
2000 of those pairs. For the zero shot class of angles, its relative output with respect to the known classes of lines, triangles, quadrilaterals and pentagons should be $(-1,1,1,1)$ respectively whereas for the known category of triangles it will be $(-1,0,1,1)$ respectively. The novelty detection percentage is
calculated as the percentage of instances in which when an unseen
class example is accurately characterized with respect to
all of the other memorized classes during testing. On the
other hand, false positive detection is the number of test
examples belonging to the previously known categories which were wrongly identified as a
different class by the memorized  samples in
the representation bank. Table \ref{dnn} and \ref{cnn} show the
results with varying threshold level $\gamma$.

\begin{table}[htbp]
    \caption{Detection as well as characterization of true and falsely detected novel geometrical shapes with a DNN learner  }\label{dnn}
    \begin{tabularx}{3.5in}{XXX}
        \hline
      Threshold Value & Novelty Detection rate & False Positive Rate \\
      \hline
      0.05 & 89.50\% & 21.70\% \\
      0.10 & 84.78\% & 15.70\% \\
      0.15 & 78.90\% & 11.10\% \\
      0.20 & 73.25\% & 7.12\% \\
      0.25 & 66.32\% & 4.98\% \\
      0.30 & 64.30\% & 3.25\% \\

      \hline
    \end{tabularx}
\end{table}

\begin{table}[htbp]
    \caption{Detection as well as characterization of true and falsely detected novel geometrical shapes with a ConvNet learner  }\label{cnn}
    \begin{tabularx}{3.5in}{XXX}
      \hline
      Threshold Value & Novelty Detection rate & False Positive Rate \\
      \hline
      0.05 & 90.65\% & 24.60\% \\
      0.10 & 88.78\% & 18.51\% \\
      0.15 & 86.56\% & 13.91\% \\
      0.20 & 84.09\% & 9.56\% \\
      0.25 & 82.03\% & 6.69\% \\
      0.30 & 77.22\% & 4.30\% \\

      \hline
    \end{tabularx}
\end{table}

 The choice of $\gamma$ leads to a trade off between the ability to identify known patterns versus  the ability to characterize novel patterns.
 It is also analogous with the human aging process. Adults tend to increase their threshold as they grow restricting them to adapt to learn new things
 while being very prudent in identifying patterns they already know and are within their comfort zone.
 On the other hand, learning children with higher inquisitiveness maintain a lower threshold and are thus very good at analyzing newer observations. Our results have a direct congruence with human learning psychology. The threshold can be varied depending on the expectations from the practical task in hand. By having a common deep architecture to do both learning and reasoning based on structured memories, we have tried to emulate the functions of both the the neocortex and hippocampus towards a holistic processing and understanding of the data. 

\subsection{CIFAR10 Object Recognition}

\begin{figure}[htbp] 
\includegraphics[width=3.5in]{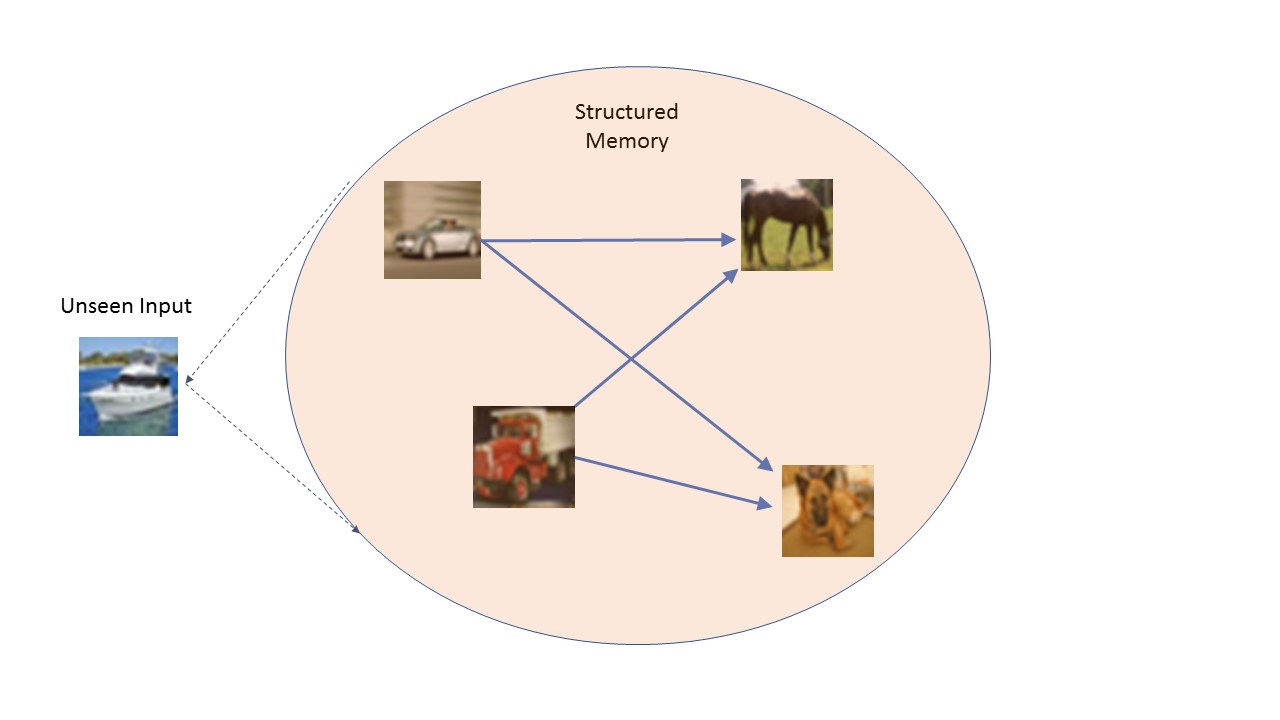}
\caption{Structured Memory formation for known CIFAR classes. Only positive edges are shown. For each edge, there exists a negative edge in the reverse direction with edge weight equals to -1. Self connections and no connections have zero weight.}
\label{struct_cifar}
\end{figure}

The proposed methodology has also been applied onto real world datasets like the CIFAR-10. Object recognition is considered as the objective for the primary learner and a ConvNet as discussed earlier is considered for it. There are many relative attributes like size, speed and other characteristics that can define the primary object classes in the input images. The representation bank is structured by the comparative learner using the relative concept of whether one is more \emph{lively} than the other. The exact question can be framed as - \emph{Is the left input image (+1) a living object as compared to the right or vice versa (-1) or do they both belong to same status (0) ?}. For example, trucks and birds will get a score of $-1$ whereas dogs and birds will get a score of 0. Just like the previous task on synthetic dataset, Fig. \ref{struct_cifar} shows the four classes that were chosen as the known object classes for training- automobiles, trucks, horses and dogs. The first two are non-living whereas the last two are living objects. A simple LeNet style ConvNet, as mentioned in the previous section,  
could attain upto $85.7\%$ recognition accuracy on classifying these four known classes. Now, images belonging to the 'ships' category were introduced as zero shot object examples which the network has never seen. So, with respect to the known classes of automobiles, trucks, horses and dogs, the correct relationship should be $0,0,-1, -1$ respectively. As can be seen in Table \ref{uncertain}, almost half of these examples fail the uncertainty test, i.e. these zero shot examples get easily identified with very high confidence as one of the known classes by just the primary learner ConvNet. The threshold for uncertainty is chosen as the mean of the softmax activations for the correct class. This suggests that uncertainty alone cannot be considered sufficient for outlier or novelty detection. On the other hand, using the comparative learner helps to identify the right relationship of the unseen ship class to the known classes at an accuracy of $85.2\%$.  Unlike the experiments with the synthetic polygon images in the previous section, the comparative learner was trained as a classifier and class wise majority voting was used to pool the decisions of each member of the representation bank. Table \ref{uncertain} also shows that the results are slightly better if the representation bank is created using samples just after the last pooling layer as compared to the fully connected layers. It is tricky to say which layer is the best to sample from since the very last layers in the primary learner are usually the most discriminative ones but lose a lot of intrinsic information regarding the original input distribution unconditioned on its corresponding classes. Lower layers, on the other hand, are much less discriminative and are usually generic feature extractors like edge and blob detectors.

\begin{table}[htbp]
    \caption{Accuracy in characterization of unseen class (ships)  }\label{uncertain}
    \begin{tabularx}{3.5in}{XX}
        \hline
				Type & Accuracy of correct characterization  \\
      \hline
      \text{Uncertainty based detection} & 54.9\%  \\
      \text{Our model with reps at FC-1} & 82.9\%  \\
      \text{Our model with reps at FLATTEN0} & 85.2\%  \\

      \hline
    \end{tabularx}
\end{table}

\begin{figure}[htbp] 
\includegraphics[width=3.5in]{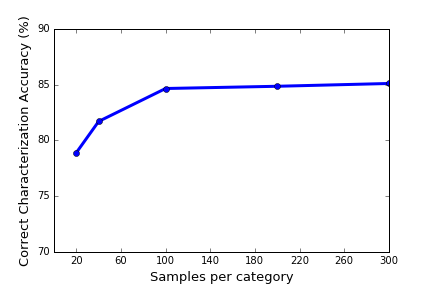}
\caption{Variation of zero shot characterization accuracy with respect to the number of samples per known category}
\label{sample}
\end{figure}

The training data had 5000 instances per class and it is interesting to see how the performance varies when we increase the sampling per class. Fig. \ref{sample} shows the variation with respect to the samples per category stored in the representation bank where it is shown that the accuracy starts saturating as $M$ increases. Thus with a constant $M$, it reinstates the claim that the number of representations, and hence the time complexity, required for training the structured memory is $\mathbf{O}(C^2)$ instead of being $\mathbf{O}(N^2)$.

\section{Conclusion}

Novel image characterization is just one application domain of the proposed system. Several other tasks can also be put into this framework. It can  be
applied on natural language understanding where instead of the sole
intent of document categorization, we can obtain abstract feature
representation where both grouping and comparative relationships
can be performed at ease. This can help detect and appreciate an
novel genre of writing style which might have been absent in the training set, for example bizarro fiction which uses elements of satire, absurdism and other categories. Our current experiment shows that even
discriminatively trained primary learners like ConvNets, if properly
regularized, can also create memory worthy features that can
suitably create a comparative representation bank. An explicit structural memory based deep learning framework has
been proposed which can be tuned as per the practitioner's needs
in regards to the trade off between better recognition of known
concepts or better creativity for understanding novelty. One single system can now perform recognition of known instance categories as well as characterization of unknown ones.

\ifCLASSOPTIONcaptionsoff
  \newpage
\fi



\bibliographystyle{IEEEtran}
\bibliography{sample}


\end{document}